\documentclass[a4paper,UKenglish,cleveref, autoref, thm-restate]{lipics-v2021}

\usepackage{kotex}
\usepackage{xcolor}

\title{Chain of Grounded Objectives: Concise  Goal-oriented Prompting for Code Generation} 

\author{Sangyeop Yeo}{Electronics and Telecommunications Research Institute, Korea}{sangyeop@etri.re.kr}{https://orcid.org/0009-0007-3413-0240}{}
\author{Seung-Won Hwang}{Department of Computer Science and Engineering, Seoul National University, Korea}{seungwonh@snu.ac.kr}{https://orcid.org/0000-0003-0782-0661}{}
\author{Yu-Seung Ma\footnote{corresponding author}}{Electronics and Telecommunications Research Institute, Korea}{ysma@etri.re.kr}{https://orcid.org/0000-0002-4168-5515}{}

\authorrunning{J. Open Access and J.\,R. Public} 

\Copyright{Jane Open Access and Joan R. Public} 



\category{} 

\funding{This work was supported by Institute of Information \& communications Technology Planning \& Evaluation (IITP) grant funded by the Korea government (MSIT) (No.2022-0-00995, Automated reliable source code generation from natural language descriptions)}

\acknowledgements{We appreciate the anonymous reviewers' constructive feedback that improved our paper.}

\EventEditors{Jonathan Aldrich and Alexandra Silva}
\EventNoEds{2}
\EventLongTitle{39th European Conference on Object-Oriented Programming (ECOOP 2025)}
\EventShortTitle{ECOOP 2025}
\EventAcronym{ECOOP}
\EventYear{2025}
\EventDate{June 30--July 2, 2025}
\EventLocation{Bergen, Norway}
\EventLogo{}
\SeriesVolume{333}
\ArticleNo{42}

\begin{document}
{
\maketitle

\begin{abstract}
The use of Large Language Models (LLMs) for code generation has gained significant attention in recent years. Existing methods often aim to improve the quality of generated code by incorporating additional contextual information or guidance into input prompts. Many of these approaches adopt process-oriented reasoning strategies, mimicking human-like step-by-step thinking; however, they may not always align with the structured nature of programming languages. This paper introduces {\bf Chain of Grounded Objectives (CGO)}, a concise goal-oriented prompting approach that embeds functional objectives into prompts to enhance code generation. By focusing on precisely defined objectives rather than explicit procedural steps, CGO aligns more naturally with programming tasks while retaining flexibility. Empirical evaluations on HumanEval, MBPP, their extended versions, and LiveCodeBench show that CGO achieves accuracy comparable to or better than existing methods while using fewer tokens, making it a more efficient approach to LLM-based code generation.
\end{abstract}

\section{Introduction}
\label{sec:typesetting-summary}
The advancement of Large Language Models (LLMs) has significantly transformed software development, particularly in natural language-driven code generation~\cite{zhu2024deepseek, hui2024qwen2}. These models enable developers to generate code from textual descriptions, reducing manual effort and expanding accessibility for non-experts. However, fully leveraging their potential requires addressing challenges related to output accuracy, motivating researchers to explore more structured and detailed prompts. In response, recent research has focused on  enhancing the input to LLMs by incorporating additional context or guidance, and these methods can be classified according to the type of supplementary information provided:

First, \textbf{example-based guidance} approaches involve supplying sample codes or problem-solution examples to guide LLMs; with few-shot prompting~\cite{brown2020language} being a typical example. 

Second, \textbf{process-oriented reasoning},  focuses on guiding the model through the procedural steps required to solve a problem, essentially explaining "how" to reach a solution. Prompts like Chain of Thought~\cite{wei2022chain} and self-planning~\cite{jiang2024self} exemplify this approach by decomposing the problem-solving process into a sequence of logical steps. Based on our preliminary experiments, while effective, process-oriented reasoning can introduce unnecessary complexity in straightforward tasks, potentially leading to increased inference time and token consumption.

Third, \textbf{goal-oriented reasoning} focuses on specifying "what" needs to be achieved rather than prescribing "how" to achieve it. This includes prompting with input-output examples, specifications, or constraints, leaving the model to infer the optimal approach. Test-driven prompting~\cite{tian2023test, Mathews_2024} is a representative example: the problem goal is defined by test cases, but such methods require code execution for validation, incurring computational overhead. 

However, {\bf is complex reasoning always necessary for effective code generation?} Instead of guiding LLMs through intricate reasoning steps, could a simpler approach—one that provides only clear and direct goals—lead to better results?

To explore this, we propose {\bf Chain of Grounded Objectives (CGO)}, a concise goal-oriented reasoning approach that succinctly captures the objectives of a coding problem. CGO first prompts the LLM to analyze the given problem and autonomously generate a compact set of objectives outlining the functional requirements of the solution. These objectives are then incorporated as supplemental input to guide the final code generation step. Notably, they are expressed in a style similar to code comments, a format LLMs have extensively encountered during training. By presenting objectives in this grounded manner, CGO enhances compatibility with programming languages while prioritizing computational interpretability. Furthermore, since CGO introduces minimal additional information, it consumes fewer tokens than verbose, process-oriented prompts, making it a more efficient alternative.

By examining CGO's effectiveness and efficiency in comparison to existing reasoning techniques, we aim to determine whether concise goal-driven guidance can serve as a more practical alternative to complex procedural reasoning in code generation.

The key contributions of this study are as follows:

\begin{itemize}
    \item \textbf{Chain of Grounded Objectives (CGO)} is introduced as a concise goal-oriented reasoning prompt technique that employs functional objectives to guide code generation.
    \item \textbf{Higher accuracy}: Experimental results on HumanEval and MBPP show that CGO outperforms existing prompting techniques.
    \item \textbf{Real-world applicability}: CGO also exhibits competitive performance on LiveCode Bench, a benchmark reflecting more realistic coding scenarios.
    \item \textbf{Efficiency}: Despite requiring fewer tokens, CGO maintains high-quality code generation relative to other prompt techniques.
\end{itemize}

\section{Related Work}

\subsection{Code Generation Using LLMs}
The development of pretrained language models has significantly influenced research in code generation. Models such as CodeT5~\cite{wang2021codet5} and UnixCoder~\cite{guo2022unixcoder}, which learn from both textual and code data, have demonstrated strong performance across various code generation and understanding tasks. Furthermore, models like Codex, CodeGen, AlphaCode, and CodeGeeX have shown that incremental increases in model size can lead to improved performance, highlighting the importance of scaling model parameters in code generation models~\cite{chen2021evaluatinglargelanguagemodels,nijkamp2022codegen,Li_2022,zheng2024codegeexpretrainedmodelcode}.

Recently, open-source models have also made significant progress. Models such as LLaMA and Mistral~\cite{touvron2023llamaopenefficientfoundation, jiang2023mistral7b} provide researchers with accessible yet high-performing alternatives, while CodeLLaMA~\cite{touvron2023llama2openfoundation}, specifically designed for code generation, has achieved strong performance in code-related tasks. Moreover, the emergence of small open-weight models like Deepseek and Qwen has further improved the efficiency of code generation models. Qwen~\cite{hui2024qwen2} has released models optimized not only for general natural language processing (NLP) tasks but also for code generation, demonstrating performance comparable to some large proprietary models. Deepseek introduced various architectures, including Deepseek-R1~\cite{guo2025deepseek}, which enhances Deepseek-V3 with efficient reinforce learning, excelling in benchmarks and competitive programming.

Additionally, proprietary models such as the GPT series, Gemini, and Claude continue to achieve state-of-the-art performance in code generation, demonstrating ongoing advancements in code understanding and generation capabilities~\cite{achiam2023gpt,team2023gemini}.

\subsection{Prompting Strategies for Code Generation}
The simplest way to generate code with LLMs is to input the coding problem directly and let the model produce a solution. However, as effective prompting has been shown to improve LLM performance, various studies have explored prompting techniques for code generation.

Example-based guidance approaches, such as few-shot prompting~\cite{brown2020language}, provide example problem-solution pairs but have shown limited impact on code quality, according to our preliminary experiments.
Process-oriented reasoning methods decompose the problem-solving process into a sequence of manageable steps. CodeCoT adapts CoT for code generation by refining intermediate steps to align with coding tasks~\cite{huang2023codecot}, Structured Chain of Thought~(SCoT) uses pseudocode as intermediate steps~\cite{li2023structured}.
Self-Planning~\cite{jiang2024self} prompts the model to create a detailed plan before generating code, effectively breaking complex problems into smaller, manageable subtasks.
Goal-oriented reasoning includes methods like Test-Case-Driven Chain of Thought (TCoT)~\cite{tian2023test}. TCoT focuses on defining goals in the form of test cases to validate generated code through execution and iterative refinement. However, its reliance on execution-based feedback loop incurs computational costs.

\section{Proposed Technique: CGO Prompting}

In this study, we propose Chain of Grounded Objectives~(CGO) prompting, a concise goal-oriented reasoning approach designed to enhance the performance of LLMs in automated code generation tasks. The core idea behind CGO is to provide functional objectives as additional input to guide the LLM. These objectives are presented in the form of comment-level representations that are likely familiar to the LLM based on its training data, such as structured natural language descriptions commonly found alongside code.

\subsection{Motivation}

CGO is grounded in the principle that learning-aligned prompting has been shown to be effective~\cite{trivedi2025align}, and that providing flexible guidance rather than overly specific context can lead to improved generation performance~\cite{agarwal2024promptwizard}. Specifically, we adopt comment-style objectives based on the intuition that these resemble structured explanations commonly found in the training data of LLMs. Such structured explanations, including comments and documentation, generally describe intended functionality in a way that abstracts implementation details.
By incorporating such familiar patterns, we expect that CGO prompting:

\begin{figure*}[t]
  \includegraphics[width=\textwidth]{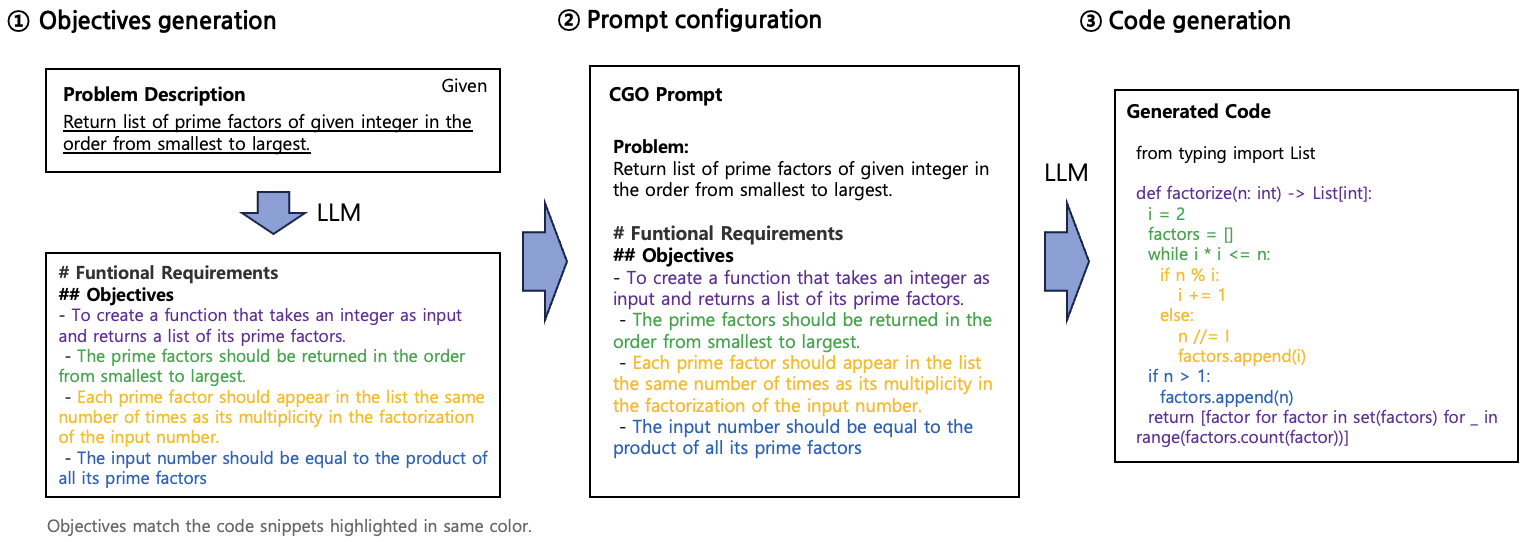}
  \caption{CGO's Workflow.}
  \label{fig:cgo_example}
\end{figure*}

\begin{figure*}[t]
  \includegraphics[width=\textwidth]{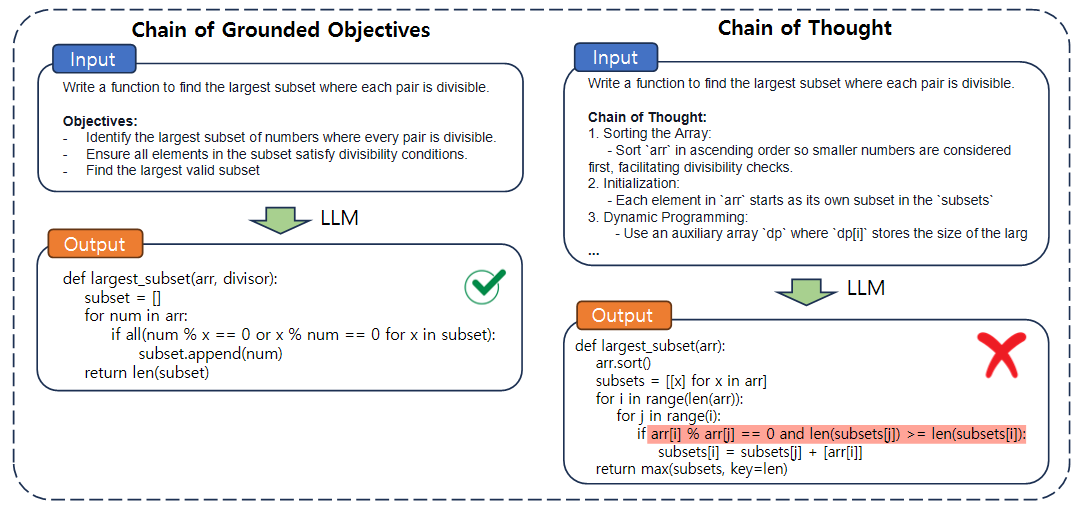}
  \caption{Code example generated with the Chain of Grounded Objectives and Chain of Thought.}
  \label{fig:cgo_example_2}
\end{figure*}

\begin{itemize}
  \item Provides clarity to the LLM about the task requirements.
  \item Improves the model’s ability to map functional objectives to code implementations.
  \item Leverage its training familiarity with structured descriptions to improve the overall quality of code generation.
\end{itemize}

Figure~\ref{fig:cgo_example} illustrates the workflow of the CGO-based code generation process consists of two stages: (1) the `Objective Generation Stage', where the LLM generates functional objectives from the problem description, and (2) the `Code Generation Stage', where these objectives are used as additional context to guide the LLM in generating the code.

The CGO-based code generation process can be formally described as follows (detailed examples are provided in Tables~\ref{cgo_example_table}):

\subsection*{(1) Objective Generation Stage}
In the first stage, the LLM generates a set of functional objectives \( O = \{o_1, o_2, \dots, o_n\} \) based on the given problem description \( P \). 

\begin{equation}
  \label{eq:obj_gen}
  O = \text{GenObj}(P)
\end{equation}

Here, \( \text{GenObj} \) is a function that maps the problem description \( P \) to a set of objectives \( O \), where each objective \( o_i \) represents a specific sub-goal, such as a functional requirement, that the final code is expected to achieve. The LLM is guided by a specifically designed prompt, such as \textit{"Write \textbf{objectives} for the problem."} 
This step ensures that the model explicitly understands the key elements of the problem before generating code.

By framing functional objectives to reflect the developer's intent, this approach bridges natural language and code, using the LLM's reasoning to align with the developer's thought process.

\subsection*{(2) Code Generation Stage}
In the second stage, the LLM synthesizes the final code \( C \) using both the problem description \( P \) and the  objectives \( O \) as input. This process is represented by the function \( \text{GenCode} \):
\begin{equation}
  \label{eq:code_gen}
 C = \text{GenCode}(O \mid P)
\end{equation}
The notation \( O\mid P\) indicates that while the objectives \( O \) are derived from the problem \( P \), they are treated as additional context that supplements the original problem description. The function \( \text{GenCode} \) leverages this enriched context to generate code that not only solves the problem but also adheres to the specific goals outlined in \( O \). This supports the LLM in developing a clear understanding of the problem’s requirements, resulting in code that is both functionally correct and contextually appropriate. 

The entire CGO-based code generation process can be represented by combining the two stages:
\begin{equation}
  \label{eq:code_full}
   C = \text{GenCode}(\text{GenObj}(P) \mid P)
\end{equation}

\section{Evaluation}

\subsection{Experimental Setup}

\subsubsection{Benchmarks}
Experiments were conducted on five datasets: HumanEval, MBPP-sanitized, their extended versions—HumanEval+ and MBPP+—which include additional edge cases, and LiveCodeBench. 

While HumanEval and MBPP-sanitized provide high-quality coding problems for evaluating LLMs, they primarily consist of standalone function implementations, making them relatively simple benchmarks. HumanEval+ and MBPP+ improve robustness assessment by incorporating more test cases, while LiveCodeBench better reflects real-world coding challenges by requiring a deeper understanding of problem contexts and constraints. These datasets support  the evaluation of both fundamental code generation capabilities and performance in practical settings.

\begin{itemize}
    \item \textbf{HumanEval}: A benchmark of 164 coding problems introduced by OpenAI, each comprising a natural language description, example function signatures, and an average of 7.7 test cases per problem.
    \item \textbf{MBPP-sanitized}~\cite{austin2021program}:  A refined subset of MBPP, containing verified problems with a problem description, a reference solution, and three test cases for each instance.
    \item \textbf{HumanEval+ and MBPP+}~\cite{liu2024your}: Extended versions of HumanEval and MBPP with 80x and 35x more test cases, focusing on edge case robustness.
    \item \textbf{LiveCodeBench}~\cite{jain2024livecodebenchholisticcontaminationfree}  : A contamination-free and continuously evolving benchmark for evaluating LLMs on code generation. Each problem in LiveCodeBench includes a structured problem description, example inputs and outputs, and explicit constraints, making it more reflective of real-world programming tasks.
\end{itemize}

In this experiment, since MBPP-sanitized does not provide function signatures, the input prompt was adjusted to follow the {\bf HumanEval-style} for consistency by extracting function headers from the ground-truth code. Examples of this formatting are shown in Table~\ref{humaneval_style_example_table}.

\subsubsection{Baselines}
\label{sec:baseline}
We compared the proposed method with the following baseline approaches (detailed examples are provided in Tables~\ref{few-shot-example} to~\ref{pseudo-example}):

\begin{itemize}
    \item \textbf{Direct prompting}: Generates code solely from the problem description.
    \item \textbf{Few-shot}: Includes example problems and solutions to guide the model.
    \item \textbf{CodeCoT}: Utilizes Chain of Thought (CoT) with a one-shot example to generate intermediate steps and final code in two stages~\cite{huang2023codecot}.
    \item \textbf{Zero-shot CoT}: A prompt that generates a CoT Reasoning process to solve the given problem without providing example CoT and incorporates it into code generation~\cite{kojima2023largelanguagemodelszeroshot}.
    \item \textbf{Self-Planning}: Extracts sub-tasks from the problem and generates a plan, which guides code generation in two stages~\cite{jiang2024self}.
    \item \textbf{Self-Pseudo}: Reflecting the design phase of the software development process, first define the problem requirements and then generate pseudocode based on them. Subsequently, use the generated pseudocode to write the final code in three stage.
\end{itemize}

\subsubsection{Performance Evaluation}
We compared the proposed CGO against six baseline reasoning methods in Section~\ref{sec:baseline} using five LLMs: LLaMA3-8B-Instruct, LLaMA3-70B-Instruct, LLaMA3.1-8B-Instruct, LLaMA3.1-70B-Instruct, and GPT-3.5-Turbo. 
For all methods, both the intermediate reasoning steps and the final code generation were performed using \textit{greedy sampling} (\textit{temperature} = 0, \textit{top-p} = 1), producing deterministic outputs at each stage. Further details of the experimental setup are provided in Appendix~\ref{sec:Expi_detail}.

Performance evaluation was conducted using \textit{pass@k}, a widely adopted metric, along with \textit{pass-ratio@n}, which provides a more fine-grained analysis of code generation performance. Further details regarding the metrics bellow are provided in  Appendix~\ref{sec:eval_detail}.

\begin{itemize} 
    \item \textit{pass@k}: The probability of generating at least one correct solution among $k$ attempts~\cite{chen2021evaluatinglargelanguagemodels}. 
    \item \textit{pass-ratio@n}: The average squared ratio of passed test cases across $n$ generated solutions, emphasizing higher accuracy, as defined in~\cite{yeo2024framework}. 
\end{itemize}

Although \textit{greedy sampling} typically produces the same output for a given prompt, minor variations may still occur due to factors such as floating-point precision errors and non-deterministic execution environments. To account for this, we repeated the experiment 10 times per problem. As a result, we set $k=1$ for \textit{pass@k}, while for \textit{pass-ratio@n}, we used $n=10$, averaging results across multiple runs. 

\begin{figure*}[t]
  \includegraphics[width=\textwidth]{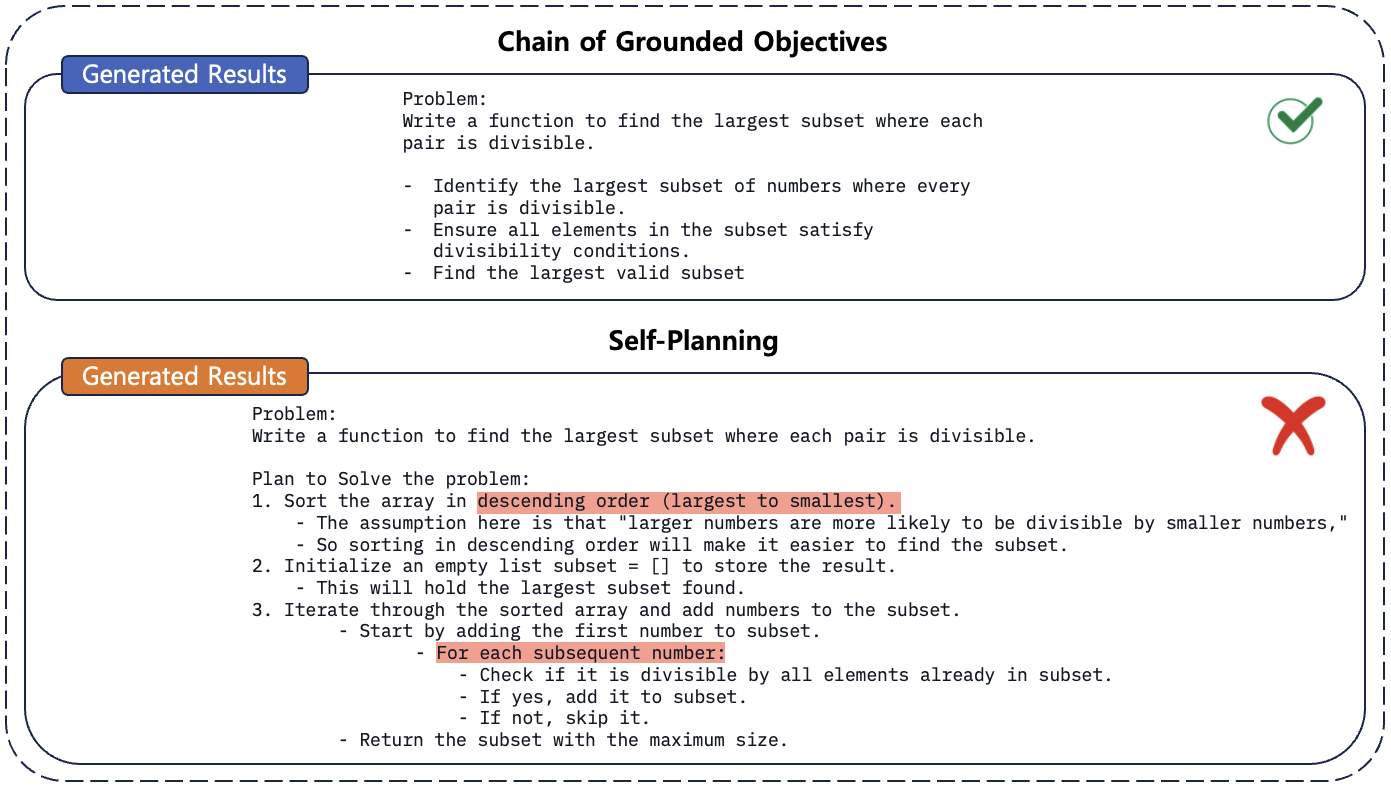}
  \caption{Context example generated with the Chain of Grounded Objectives and Self-Planning.}
  \label{fig:cgo_planning}
\end{figure*}

\subsection{RQ 1: How Does CGO Perform Compared to Baselines?}
In this section, we evaluate the correctness of generated code across four benchmarks: HumanEval, MBPP-sanitized, and their extended versions—HumanEval+ and MBPP+. Tables~\ref{Table1}, \ref{Table2}, and \ref{Table3} present the \textit{pass@1} (for $n=10$) and \textit{pass-ratio@10} scores, comparing CGO with various baseline reasoning methods using three different LLMs: LLaMA3-8B-Instruct, LLaMA3-70B-Instruct, and GPT-3.5-Turbo.

\subsubsection*{LLaMA3-8B-Instruct (Medium-scale LLM)}

Table~\ref{Table1} reports the results for LLaMA3-8B-Instruct, showing the accuracy of code generated using nine different methods: Direct prompting, 1-shot, 2-shot, 3-shot, CodeCoT, zero-shot CoT, self-planning, self-pseudo, and CGO. The results demonstrate that CGO outperformed all baseline reasoning methods across all benchmarks, achieving the highest \textit{pass@1} and \textit{pass-ratio@10} scores. 

Interestingly, despite its widespread use in LLM prompting, few-shot prompting yielded surprisingly suboptimal performance, with inconsistent results across different benchmarks. While it showed  improvements in MBPP-sanitized and MBPP+, it underperformed compared to direct prompting on the HumanEval benchmarks. Overall, it did not demonstrate a clear advantage over other reasoning methods, suggesting that few-shot prompting may be less effective for code generation than anticipated.


CodeCoT improved over few-shot prompting, demonstrating the benefits of leveraging CoT, but still fell 4.9 points behind CGO in \textit{pass@1} (57.5 vs. 62.4) and 4.1 points in \textit{pass-ratio@10} (61.3 vs. 65.4) on HumanEval.  Across the evaluated benchmarks, Zero-shot CoT often outperformed CodeCoT; however, it still lagged behind CGO, suggesting that leveraging the model's inherent reasoning abilities may be more effective than manually constructing CoT examples for code generation.

While CGO and CoT share the general principle of decomposition, they differ in their underlying reasoning processes. CoT emulates a human-like, step-by-step reasoning, whereas CGO adopts a machine-oriented reasoning approach. This mechanistic approach facilitates a more direct alignment with automated code-generation tasks. 
Figure~\ref{fig:cgo_example_2} presents an example of the intermediate outputs generated by CGO (objectives) and CoT (step-by-step reasoning), clearly demonstrating that CGO generates more concise intermediate results. While CoT’s detailed, explanatory style is beneficial to some extent, Figure~\ref{fig:cgo_planning} shows that its emphasis on human-readable explanations introduces unnecessary verbosity, diverting focus from capturing essential functional objectives for code generation.

\begin{table*}
  \renewcommand{\arraystretch}{1.0}
  \centering
  \resizebox{\textwidth}{!}{%
  \begin{tabular}{lcccccccc}
    \hline
    \textbf{Prompts} & \multicolumn{2}{c}{\textbf{HumanEval}} & \multicolumn{2}{c}{\textbf{MBPP-sanitized}} & \multicolumn{2}{c}{\textbf{HumanEval+}} & \multicolumn{2}{c}{\textbf{MBPP+}} \\
    \cline{2-3} \cline{4-5} \cline{6-7} \cline{8-9}
                     & $pass$@$1$ & $pass$-$ratio$@$10$ & $pass$@$1$ & $pass$-$ratio$@$10$ & $pass$@$1$ & $pass$-$ratio$@$10$ & $pass$@$1$ & $pass$-$ratio$@$10$ \\
    \hline
    Direct prompting & 54.4    & 57.3          & 53.5    & 56.7          & 48.5   & 49.9          & 43.2    & 46.6          \\
    1-shot           & 47.8    & 50.8          & 60.4    & 61.4          & 42.9   & 43.6          & 50.3    & 51.1          \\
    2-shot           & 47.2    & 50.5          & 64.4    & 65.4          & 42.9    & 43.7         & 54.5    & 55.1          \\
    3-shot           & 45.4    & 48.7          & 64.8    & 65.4          & 41.5    & 42.1         & 54.9    & 55.6          \\
    CodeCoT          & 57.5    & 61.3          & 63.7    & 64.6          & 51.1   & 52.2          & 52.3    & 53.3          \\
    Zero-shot CoT    & 59.8    & 63.4          & 63.8    & 64.7          & 53.7   & 54.3          & 54.0    & 55.6          \\
    Self-Planning    & 60.1    & 63.8          & 67.9    & \textbf{68.8} & 52.8   & 53.5          & 56.9    & 57.4          \\
    Self-Pseudo      & 52.4    & 55.2          & 48.4    & 49.2          & 45.7   & 46.6          & 39.7    & 40.4          \\
    CGO              & \textbf{62.4}    & \textbf{65.4}          & \textbf{68.1}    & \textbf{68.8}          & \textbf{56.2}   & \textbf{56.9}          & \textbf{57.9}    & \textbf{58.4}          \\
    \hline
  \end{tabular}%
  }
  \caption{\label{Table1}
    Comparison of the CGO prompt with various baselines using the LLaMA3-8B-Instruct across four benchmarks (HumanEval, HumanEval+, MBPP-sanitized, MBPP+), evaluated using $pass$@$1 (n$=$10)$ and $pass$-$ratio$@$10$.
  }
\end{table*}

Among the evaluated baseline methods, self-planning achieved the highest performance, coming closest to CGO. While planning, like CoT, emphasizes procedural steps, it places less emphasis on strict step-by-step sequencing, which may contribute to its stronger performance.

Self-Pseudo follows a three-stage reasoning process, sequentially generating requirements, pseudo-code, and then the final implementation, aligning with typical software development procedures.
However, its effectiveness was limited with LLaMA3-8B-Instruct, consistently recording lower scores than direct prompting (DP) across all four benchmarks. This suggests that incorporating software development process elements into code generation is less effective for an 8B-scale model, likely due to its limited capacity to process complex contextual dependencies.

Overall, CGO consistently outperformed all other methods, demonstrating that its concise, goal-oriented reasoning approach effectively enhances code generation accuracy across diverse benchmarks.

\begin{table*}
  \renewcommand{\arraystretch}{1.0}
  \centering
  \resizebox{\textwidth}{!}{%
  \begin{tabular}{lcccccccc}
    \hline
    \textbf{Prompts} & \multicolumn{2}{c}{\textbf{HumanEval}} & \multicolumn{2}{c}{\textbf{MBPP-sanitized}} & \multicolumn{2}{c}{\textbf{HumanEval+}} & \multicolumn{2}{c}{\textbf{MBPP+}} \\
    \cline{2-3} \cline{4-5} \cline{6-7} \cline{8-9}
                     & $pass$@$1$ & $pass$-$ratio$@$10$ & $pass$@$1$ & $pass$-$ratio$@$10$ & $pass$@$1$ & $pass$-$ratio$@$10$ & $pass$@$1$ & $pass$-$ratio$@$10$ \\
    \hline
    Direct prompting & 76.9    & 80.2          & 72.7    & 74.0          & 71.3   & 72.6          & 60.1    & 63.1          \\
    1-shot           & 73.0    & 75.6          & 75.3    & 75.7          & 68.4   & 69.5          & 63.9    & 64.6          \\
    2-shot           & 73.9    & 77.3          & 76.4    & 76.8          & 70.0   & 70.9          & 65.0    & 65.8          \\
    3-shot           & 74.4    & 77.7          & 73.8    & 74.3          & 70.2   & 71.3          & 63.1    & 64.0          \\
    CodeCoT          & 76.2    & 79.6          & 82.3    & 82.9          & 68.2   & 69.8          & 69.7    & 70.7          \\
    Zero-shot CoT    & 79.3    & 82.8          & 80.6    & 81.1          & 74.4   & 74.9          & 69.0    & 69.8          \\
    Self-Planning    & 79.5    & 82.3          & 81.5    & 82.0          & 71.8   & 72.7          & 67.4    & 68.4          \\
    Self-Pseudo      & 79.9    & 81.7          & 82.3    & 82.8          & 73.8   & 75.0          & 69.3    & 70.6          \\
    CGO              & \textbf{80.3}    & \textbf{83.2}          & \textbf{83.3}    & \textbf{83.7}          & \textbf{74.6}   & \textbf{75.7}          & \textbf{70.4}    & \textbf{71.1}          \\
    \hline
  \end{tabular}%
  }
  \caption{\label{Table2}
    Comparison of the CGO prompt with various baselines using the LLaMA3-70B-Instruct across four benchmarks (HumanEval, HumanEval+, MBPP-sanitized, MBPP+), evaluated using $pass$@$1 (n$=$10)$ and $pass$-$ratio$@$10$.
  }
\end{table*}

\begin{table*}
  \renewcommand{\arraystretch}{1.0}
  \centering
  \resizebox{\textwidth}{!}{%
  \begin{tabular}{lcccccccc}
    \hline
    \textbf{Prompts} & \multicolumn{2}{c}{\textbf{HumanEval}} & \multicolumn{2}{c}{\textbf{MBPP-sanitized}} & \multicolumn{2}{c}{\textbf{HumanEval+}} & \multicolumn{2}{c}{\textbf{MBPP+}} \\
    \cline{2-3} \cline{4-5} \cline{6-7} \cline{8-9}
                     & $pass$@$1$ & $pass$-$ratio$@$10$ & $pass$@$1$ & $pass$-$ratio$@$10$ & $pass$@$1$ & $pass$-$ratio$@$10$ & $pass$@$1$ & $pass$-$ratio$@$10$ \\
    \hline
    Direct prompting & 71.6    & 75.3          & 84.3    & 84.8          & 67.7   & 68.5          & 71.2    & 71.2          \\
    1-shot           & 70.4    & 74.8          & 84.9    & 85.3          & 65.5   & 65.8          & 73.3    & 73.8          \\
    2-shot           & 72.0    & 73.9          & 83.6    & 84.1          & 67.7   & 68.0          & 73.3    & 73.7          \\
    3-shot           & 64.6    & 66.9          & 85.2    & 85.7          & 61.6   & 61.6          & 73.3    & 73.7          \\
    CodeCoT          & 74.2    & 77.2          & 85.6    & 86.3          & 67.8   & 68.6          & 72.2    & 72.6          \\
    Zero-shot CoT    & 72.0    & 74.6          & 85.2    & 85.7          & 67.1   & 67.6          & 71.7    & 72.5          \\
    Self-Planning    & 72.7    & 76.1          & 82.6    & 83.4          & 67.3   & 67.7          & 68.1    & 68.3          \\
    Self-Pseudo      & \textbf{75.4}    & 77.1          & 85.4    & 85.5          & \textbf{69.1}   & 69.0          & 72.2    & 73.1          \\
    CGO              & 74.6    & \textbf{77.4}          & \textbf{86.0}    & \textbf{86.5}          & 68.5   & \textbf{69.1}          & \textbf{73.7}    & \textbf{74.0}          \\
    \hline
  \end{tabular}%
  }
  \caption{\label{Table3}
    Comparison of the CGO prompt with various baselines using the GPT-3.5-Turbo across four benchmarks (HumanEval, HumanEval+, MBPP-sanitized, MBPP+), evaluated using $pass$@$1 (n$=$10)$ and $pass$-$ratio$@$10$.
  }
\end{table*}

\subsubsection*{LLaMA3-70B-Instruct and  GPT-3.5-turbo (Large-scale LLMs)}

Tables~\ref{Table2} and \ref{Table3} show the results for LLaMA3-70B-Instruct and GPT-3.5-Turbo, both of which have greater computational capacity than LLaMA3-8B-Instruct. With these larger models, the accuracy of all methods improved compared to LLaMA3-8B-Instruct, while the overall performance trends remained similar.

As in the previous results, CGO generally outperformed other methods across benchmarks, with self-planning achieving strong performance among baselines. Zero-shot CoT continued to perform better than CodeCoT in HumanEval, but unlike in LLaMA3-8B-Instruct, CodeCoT outperformed Zero-shot CoT in some MBPP-sanitized benchmarks, suggesting that its effectiveness may depend on the characteristics of the benchmark.

A notable difference from previous results is the performance of self-pseudo, which showed improvement with larger models. While it performed worse than direct prompting in LLaMA3-8B-Instruct, it achieved better results in HumanEval with larger models, in some cases approaching CGO. This suggests that larger models are better able to integrate multi-step reasoning, making the use of pseudo-code more effective. However, CGO still outperformed Self-Pseudo in terms of \textit{pass-ratio@10}, indicating that while generating pseudo-code can be beneficial, CGO remains a more efficient and reliable approach.

Overall, CGO demonstrated strong performance across different model sizes, showing its effectiveness as a concise, goal-oriented reasoning approach for code generation.

\subsection{RQ 2. Does CGO Effective in Real-World Programming Task?}

To evaluate CGO's effectiveness in real-world programming scenarios, we compared its performance with baseline reasoning-based prompting techniques on LiveCodeBench. Table~\ref{Table4} presents the results based on \textit{pass@1} and \textit{pass-ratio@10} across LLaMA3.1-8B-Instruct, LLaMA3.1-70B-Instruct, and GPT-3.5-Turbo.

In this experiment, few-Shot prompting was excluded due to the longer problem descriptions in LiveCodeBench. Including additional examples for few-Shot prompting would exceed the context length limit of LLaMA3.1 models, making it impractical for evaluation.

Experimental results show that while CodeCoT outperformed Direct Prompting (DP), its performance improvement was not particularly significant compared to other baseline methods. This observation became more pronounced as the model size increased. These findings suggest that as the diversity of problem increases, incorporating appropriate examples into the code generation process becomes increasingly important. 
\newpage
In contrast, zero-shot CoT demonstrated greater performance improvements compared to CodeCoT, and this tendency was consistently observed across all three models. This suggests that as problem complexity and diversity increase, leveraging the model's inherent capabilities to construct a Chain of Thought (CoT) reasoning process before code generation can be a more effective approach.

Conversely, self-planning recorded the highest performance on LLaMA3.1-8B-Instruct, achieving 30.0 in \textit{pass@1} (30.0 vs. 28.0) and 38.1 in \textit{pass-ratio@10} (38.1 vs. 34.7), outperforming CGO by 2.0 and 4.7 points. However, on LLaMA3.1-70B-Instruct and GPT-3.5-Turbo, Self-Planning showed lower performance compared to CGO.

\begin{table*}
  \renewcommand{\arraystretch}{1.0}
  \centering
  \resizebox{\textwidth}{!}{%
  \begin{tabular}{lcccccc}
    \hline
    \textbf{Prompts} & \multicolumn{2}{c}{\textbf{LLaMA3.1-8B-Instruct}} & \multicolumn{2}{c}{\textbf{LLaMA3.1-70B-Instruct}} & \multicolumn{2}{c}{\textbf{GPT-3.5-turbo}} \\
    \cline{2-3} \cline{4-5} \cline{6-7}
                     & $pass$@$1$ & $pass$-$ratio$@$10$ & $pass$@$1$ & $pass$-$ratio$@$10$ & $pass$@$1$ & $pass$-$ratio$@$10$ \\
    \hline
    Direct Prompting & 26.0    & 32.1          & 41.5    & 48.8          & 35.5   & 44.1          \\
    CodeCoT          & 28.0    & 36.1          & 44.7    & 52.4          & 35.7   & 43.6          \\
    Zero-shot CoT    & 29.0    & 36.6          & 47.0    & 54.3          & 39.0   & 48.3          \\
    Self-Planning    & \textbf{30.0}    & \textbf{38.1}          & 48.3    & 56.0          & 38.3   & 47.6          \\
    Self-Pseudo      & 25.5    & 33.0          & 48.0    & 55.7          & 39.5   & 48.1          \\
    CGO              & 28.0    & 34.7          & \textbf{49.0}    & \textbf{56.1}          & \textbf{40.75}   & \textbf{50.2} \\
    \hline
  \end{tabular}%
  }
  \caption{\label{Table4}
    Comparison of the CGO prompt with baseline methods on LiveCodeBench, using 3 LLMs. Results is evaluated based on $pass$@$1 (n$=$10)$ and $pass$-$ratio$@$10$.
  }
\end{table*}

Self-pseudo showed lower performance than DP on LLaMA3.1-8B-Instruct, consistent with the results observed in RQ 1. However, for the other two models, it demonstrated performance comparable to or higher than that of self-planning. This variation suggests that the effectiveness of the self-pseudo approach may depend on model size and capability. These findings indicate that incorporating the software development process into the code generation pipeline remains a valid approach, even in environments with diverse problem types such as LiveCodeBench. 
 
CGO showed lower performance than self-planning and zero-shot CoT when using the LLaMA3.1-8B-Instruct model. However, in experiments using LLaMA3.1-70B-Instruct and GPT-3.5-Turbo, CGO achieved the highest performance compared to the baseline prompts. Notably, in GPT-3.5-Turbo experiments, CGO recorded 40.75 in \textit{pass@1} (40.75 vs. 35.5) and 50.2 in \textit{pass-ratio@10} (50.2 vs. 44.1), 4.25 and 6.1 points higher than DP, demonstrating significant improvements.

These results suggest that CGO's goal-oriented reasoning approach is also effective in real-world problem-solving when applied to larger models such as LLaMA3.1-70B-Instruct and GPT-3.5-Turbo, rather than smaller 8B-scale models. Furthermore, it confirms CGO’s advantage over baseline prompts.

\begin{table*}
  \renewcommand{\arraystretch}{1.0}
  \centering
  \resizebox{\textwidth}{!}{%
  \begin{tabular}{lcccccc}
    \hline
    \textbf{Prompts} & \multicolumn{2}{c}{\textbf{easy}} & \multicolumn{2}{c}{\textbf{medium}} & \multicolumn{2}{c}{\textbf{hard}} \\
    \cline{2-3} \cline{4-5} \cline{6-7}
                     & $pass$@$1$ & $pass$-$ratio$@$10$ & $pass$@$1$ & $pass$-$ratio$@$10$ & $pass$@$1$ & $pass$-$ratio$@$10$ \\
    \hline
    Direct Prompting  & 60.56 & 67.83 & 22.61 & 32.73 & 20.00 & 28.00 \\
    Self-Planning     & 60.56 & 70.92 & \textbf{28.57} & 39.97 & 16.66 & 25.26 \\
    CodeCoT           & 60.56 & 67.79 & 25.00 & 34.71 & 16.66 & 22.12 \\
    Zero-shot CoT     & \textbf{65.49} & \textbf{73.78} & 27.97 & 39.02 & 17.77 & 25.99 \\
    Self-Pseudo       & 61.97 & 71.29 & \textbf{28.57} & 40.44 & 15.55 & 25.85 \\
    CGO               & \textbf{65.49} & 73.48 & \textbf{28.57} & \textbf{40.55} & \textbf{24.44} & \textbf{31.58} \\
    \hline
  \end{tabular}%
  }
  \caption{\label{Table5}
  Performance comparison of baselines and CGO across difficulty levels based on $pass$@$1$ and $pass$-$ratio$@$10$, using the GPT-3.5-Turbo.}
\end{table*}

\begin{table*}
  \renewcommand{\arraystretch}{1.0}
  \centering
  \resizebox{\textwidth}{!}{%
  \begin{tabular}{lcccccc}
    \hline
    \textbf{Prompts} & \multicolumn{2}{c}{\textbf{easy}} & \multicolumn{2}{c}{\textbf{medium}} & \multicolumn{2}{c}{\textbf{hard}} \\
    \cline{2-3} \cline{4-5} \cline{6-7}
                     & $pass$@$1$ & $pass$-$ratio$@$10$ & $pass$@$1$ & $pass$-$ratio$@$10$ & $pass$@$1$ & $pass$-$ratio$@$10$ \\
    \hline
    Direct Prompting  & 69.71 & 72.69 & 30.95 & 40.31 & 16.66 & 22.60 \\
    Self-Planning     & \textbf{80.99} & \textbf{85.33} & 38.10 & 47.09 & 15.56 & 26.52 \\
    CodeCoT           & 79.58 & 83.32 & 32.74 & 42.02 & 12.22 & 23.19 \\
    Zero-shot CoT      & 79.58 & 83.17 & 33.93 & 43.60 & \textbf{20.00} & \textbf{28.83} \\
    Self-Pseudo       & 78.16 & 82.72 & 38.28 & 47.90 & 16.88 & 28.32 \\
    CGO               & 80.22 & 82.59 & \textbf{39.88} & \textbf{49.46} & 17.78 & 26.58 \\
    \hline
  \end{tabular}%
  }
  \caption{\label{Table6}
  Performance comparison of baselines and CGO across difficulty levels based on $pass$@$1$ and $pass$-$ratio$@$10$, using the LLaMA3.1-70B-Instruct. }
\end{table*}

\subsubsection{Performance Analysis of CGO Across Difficulty Levels}
This subsection provides a more detailed analysis of Table~\ref{Table4}, examining the performance of CGO and baseline prompts across different problem difficulty levels as defined by LiveCodeBench. Table~\ref{Table5} presents the results, categorizing problems into three difficulty levels: easy, medium, and hard. The results in Table~\ref{Table5} were obtained using GPT-3.5-Turbo, while Figure~\ref{fig:two_images} provides a visual representation of these findings for better readability. 

\begin{figure}[ht]
    \centering
    \includegraphics[width=0.75\textwidth]{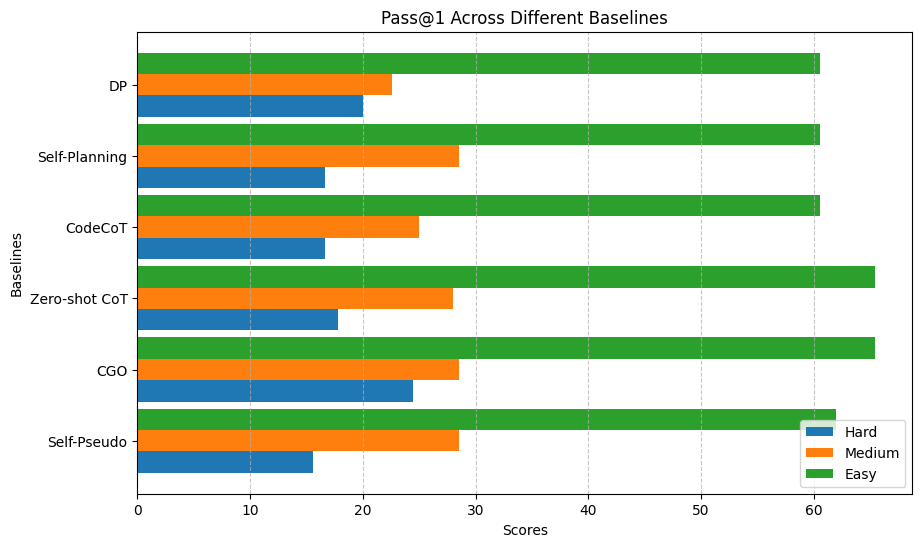}
    \vspace{2mm}  
    \includegraphics[width=0.75\textwidth]{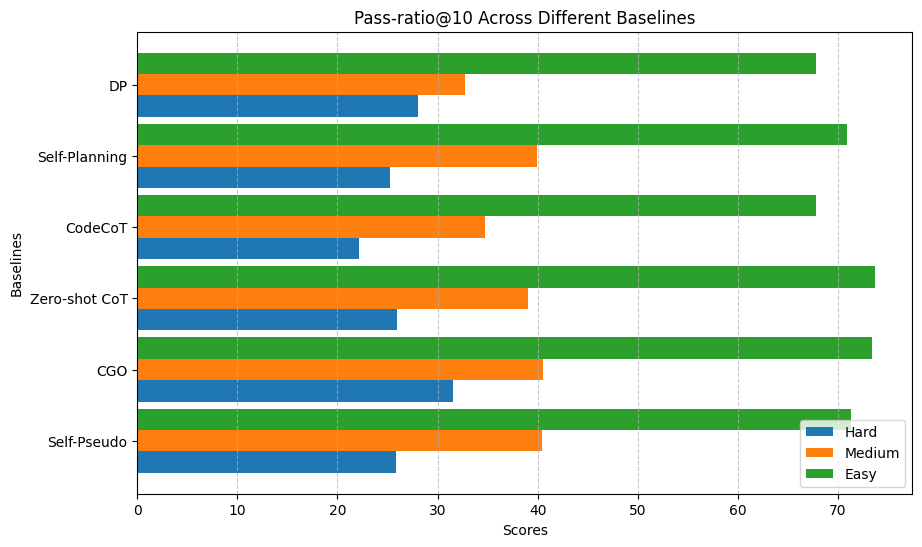}
    \caption{\textit{pass@1} and \textit{pass-ratio@10} scores by difficulty level on the GPT-3.5-turbo experiment.}
    \label{fig:two_images}
\end{figure}

Easy problems showed relatively modest performance gains across most baselines. The self-planning approach matched the baseline in \textit{pass@1} but demonstrated improvements in \textit{pass-ratio@10}. Zero-shot CoT achieved one of the highest scores among all baselines, along with CGO, significantly outperforming DP. Other prompting techniques showed performance levels similar to DP, with only minor enhancements.

As problem complexity increased, the performance gap among baselines became more pronounced. On medium-level problems, self-planning achieved 28.57 in \textit{pass@1} (28.57 vs. 22.61), outperforming DP by 4.95 points. Zero-shot CoT also demonstrated improvements over DP, reinforcing the advantages of autonomous reasoning before code generation. In contrast, CodeCoT exhibited the lowest performance, indicating the need for a more refined Chain of Thought (CoT) strategy.

Several methods, including CGO, self-planning, and self-pseudo, attained the same \textit{pass@1} score (28.57). However, in terms of \textit{pass-ratio@10}, CGO recorded the highest value (40.55), followed by self-pseudo (40.44) and self-planning (39.97). These results suggest that while different approaches may achieve similar \textit{pass@1} scores, \textit{pass-ratio@10} provides additional insight into solution quality by reflecting the proportion of generated code that successfully passes more test cases, thereby offering a more comprehensive measure of correctness.

Performance degradation was more apparent in hard problems, where most baselines underperformed compared to DP. Self-planning and CodeCoT showed lower \textit{pass@1} scores compared to DP, and this pattern was consistent in \textit{pass-ratio@10} as well. While zero-shot CoT led among the baselines, it still fell short of DP, revealing limitations in handling higher complexity tasks. self-pseudo saw the most significant performance drop, indicating its ineffectiveness in solving more challenging problems. In contrast, CGO achieved the highest performance across all baselines, with \textit{pass@1} and \textit{pass-ratio@10} scores of 24.44 and 31.58, respectively. These results represent a 4.44 and 3.58 improvement over DP, demonstrating CGO’s effectiveness not only in lower difficulty levels but also in more complex problem-solving scenarios.

Table~\ref{Table6} shows the performance of the prompting methods, including CGO, across difficulty levels using the LLaMA3.1-70B-Instruct. CGO achieved strong performance on medium-difficulty problems—which constitute the majority of the benchmark—achieving the highest scores of 39.88 for \textit{pass@1} and 49.46 for \textit{pass-ratio@10}, indicating its effectiveness in addressing problems of intermediate complexity. On easy and hard problems, although CGO did not attain the highest scores, it still ranked second in terms of \textit{pass@1}, closely behind self-planning on easy problems and zero-shot CoT on hard problems,  indicating a stable level of performance across different complexities.

The results show that CGO performs similarly to other baselines on easy, medium difficulty problems, while achieving better results on hard problems. This suggests that CGO’s objective-driven approach provides useful guidance, particularly for more complex tasks. Further studies could investigate whether integrating CGO with other reasoning approaches improves performance, particularly on hard problems where additional structured reasoning may be advantageous.

\begin{figure}[ht]
    \centering
    \includegraphics[width=0.8\textwidth]{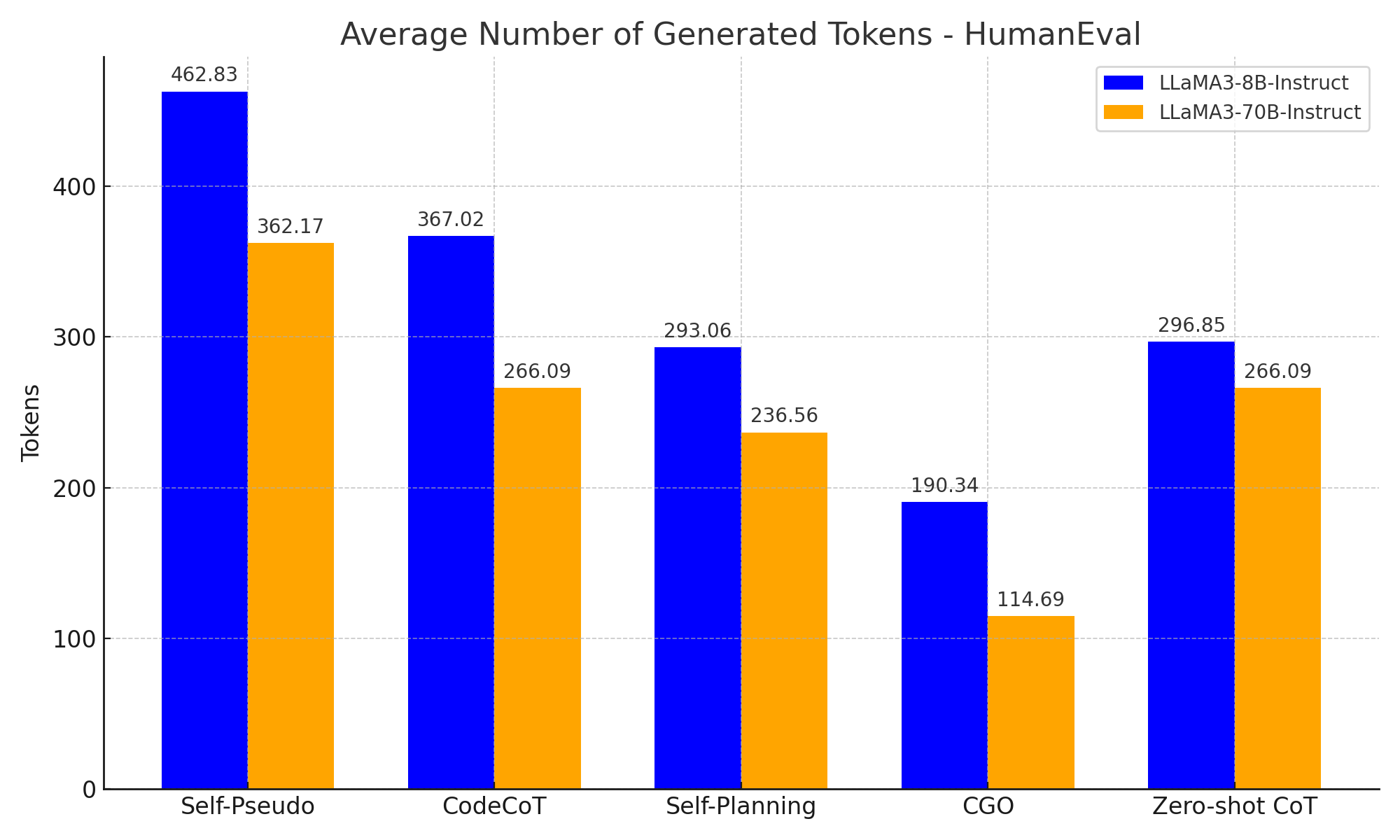}
    \vspace{2mm}  
    \includegraphics[width=0.8\textwidth]{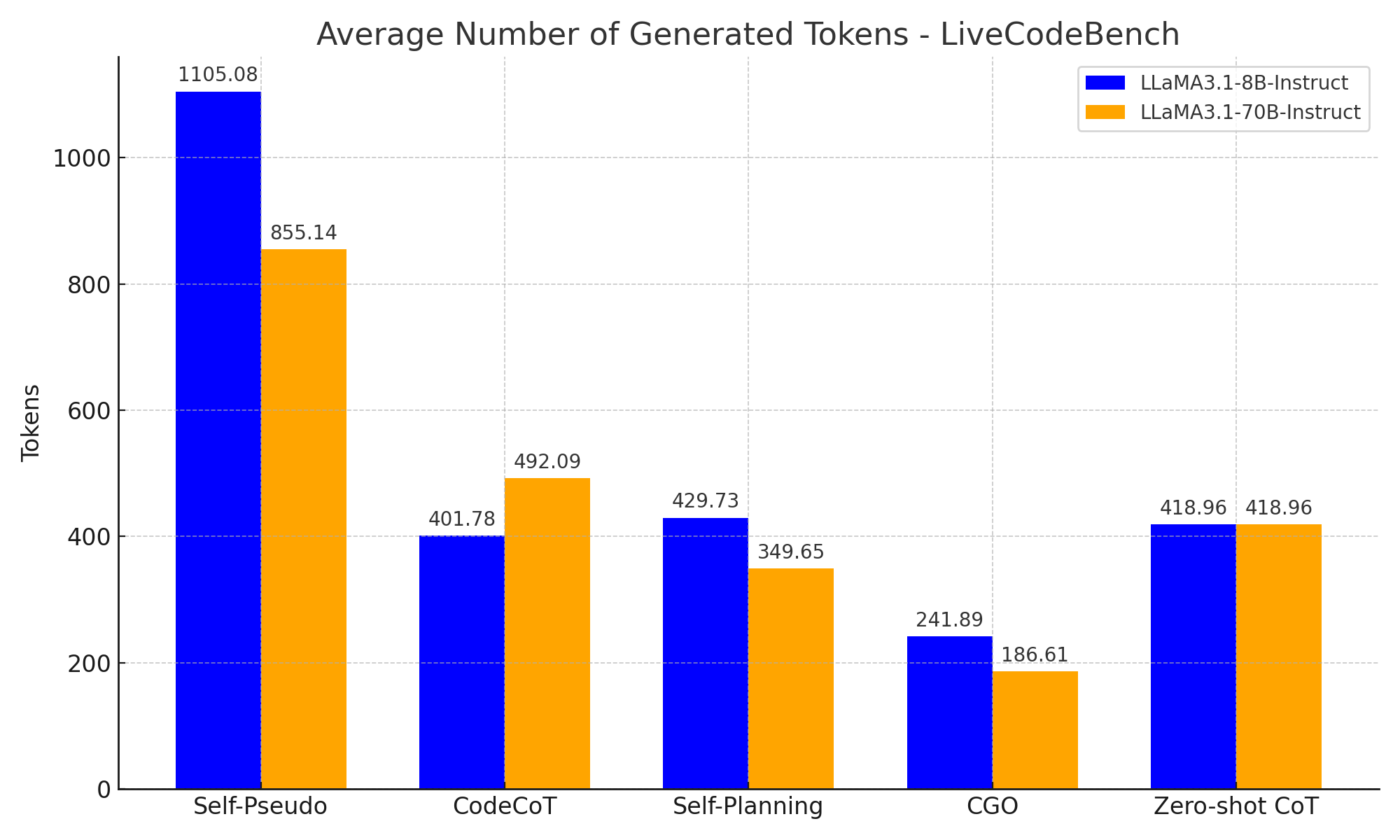}
    \caption{Average number of generated tokens by LLaMA3.1-8B-Instruct and LLaMA3.1-70B-Instruct on the HumanEval and LiveCodeBench across 5 baseline prompts (lower is better).}
    \label{fig:token_images}
\end{figure}

\begin{figure}[t]
    \centering
    \includegraphics[width=0.8\textwidth]{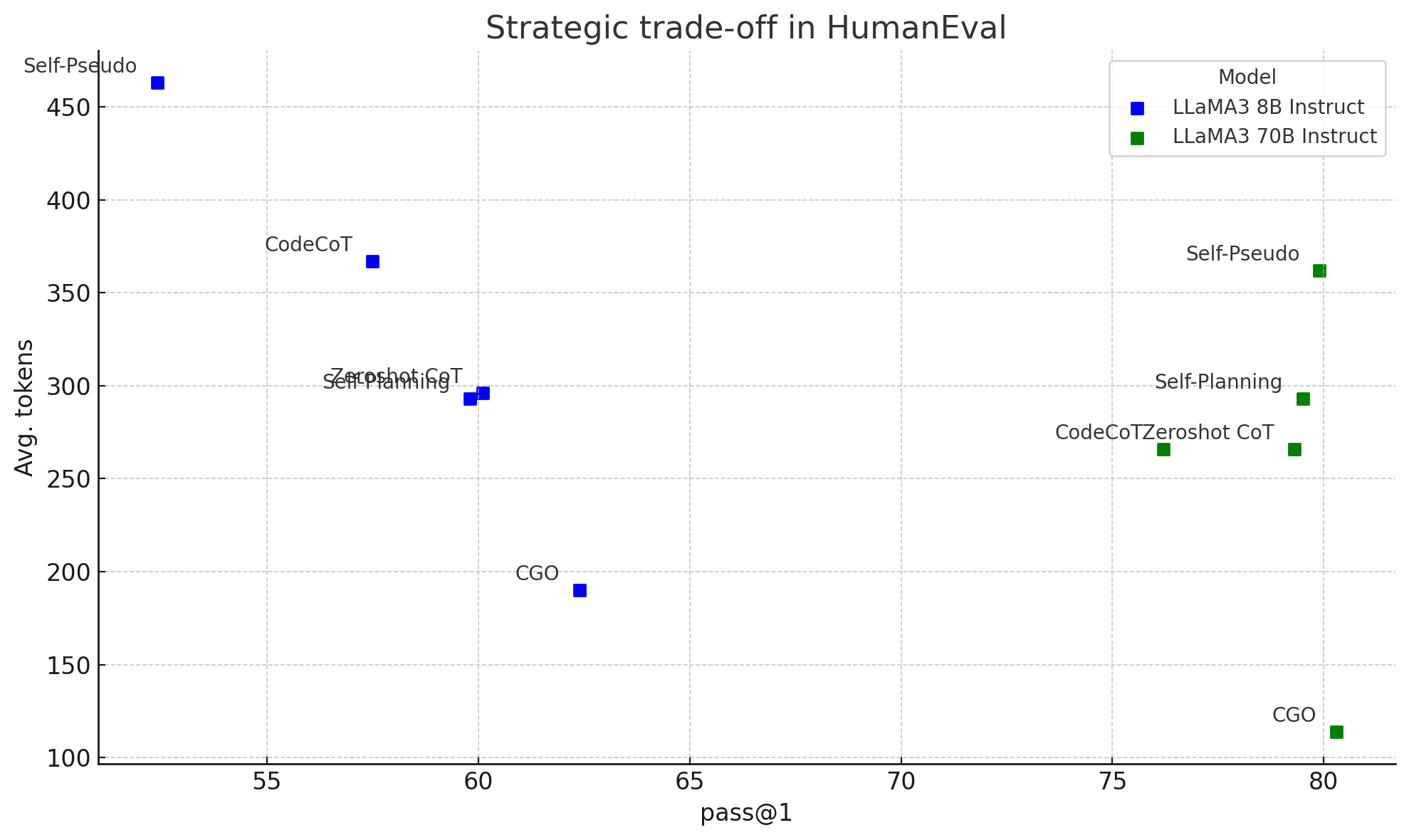}
    \vspace{2mm}  
    \includegraphics[width=0.8\textwidth]{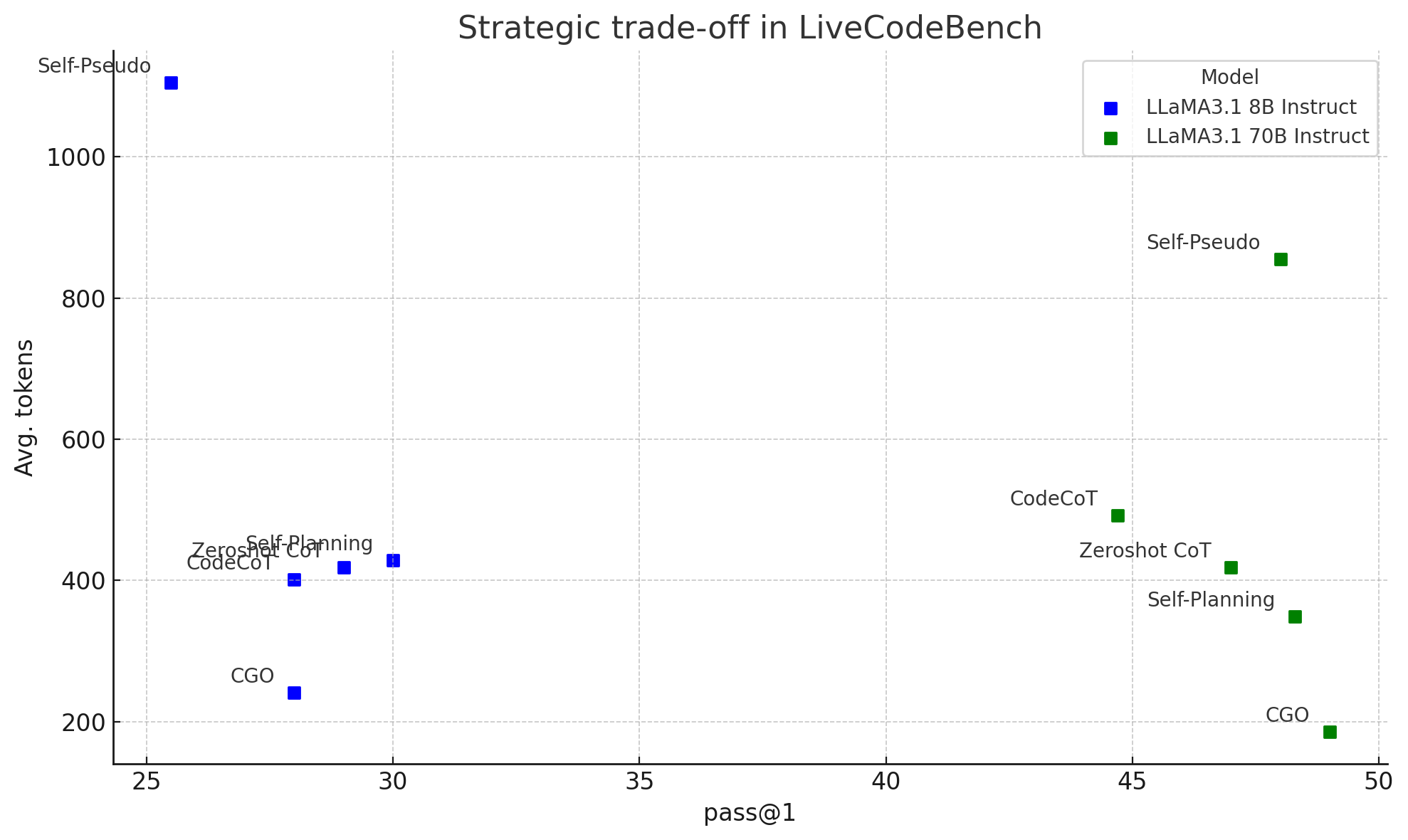}
    \caption{Strategic trade-off on the HumanEval and LiveCodeBench across 5 baseline prompts.}
    \label{fig:tradeoff_images}
\end{figure}

\subsection{RQ 3. Does CGO Resource-Efficient Compared to the Baselines?}

All prompting methods examined in this study follow a multi-step reasoning process, where the LLM first autonomously generates intermediate supplementary content (e.g., objectives, plans, chain-of-thought, or pseudo-code) before generating the final code. While this additional content helps guide the code generation process, it also increases token generation, thereby raising the computational cost of using LLMs.  Since LLM inference cost scales with the number of generated tokens, it is essential to evaluate the trade-off between enhanced guidance and resource efficiency.

In this section, we evaluate the resource efficiency of each baseline by measuring the token count of the intermediate supplementary content generated by the LLM.
Figure~\ref{fig:token_images} presents the average token count across baselines on the HumanEval and LiveCodeBench benchmarks, serving as a proxy for computational cost. 
The results show that CGO consistently generates the fewest tokens, making it the most resource-efficient approach among the compared baselines. 

On HumanEval, CGO produces 114.69 tokens on average for LLaMA3-70B-Instruct and 190.34 tokens for LLaMA3-8B-Instruct, significantly lower than other baselines. In contrast, self-pseudo requires the most tokens (462.83 for LLaMA3-8B-Instruct and 362.17 for LLaMA3-70B-Instruct), followed by CodeCoT and self-planning.

This pattern is observed on LiveCodeBench, where CGO remains the most efficient, generating 186.61 tokens for LLaMA3.1-70B-Instruct and 241.89 tokens for LLaMA3.1-8B-Instruct. self-pseudo again exhibits the highest token consumption (1105.08 and 855.14, respectively), with CodeCoT and self-planning also showing significantly higher token usage than CGO.

Interestingly, for most cases, the LLaMA3.1-8B-Instruct generates more tokens than the LLaMA3.1-70B-Instruct under the same benchmark and prompting technique. A possible explanation is that the larger LLaMA3.1-70B-Instruct, having a greater capacity to process and distill information, tends to generate more concise outputs. Consequently, it may require fewer tokens to convey the necessary details for code generation.

Overall, these findings highlight CGO as the most resource-efficient approach among the evaluated baselines. By emphasizing concise functional objectives, CGO achieves an effective balance between token usage and guidance quality, thereby minimizing computational overhead without compromising performance, as further demonstrated in Figure~\ref{fig:tradeoff_images}. Its primary advantage lies in its ability to generate only the essential contextual elements required for code generation, making it particularly well-suited for scenarios where computational efficiency is a priority.

\begin{table*}
  \renewcommand{\arraystretch}{1.0}
  \centering
  \resizebox{0.70\textwidth}{!}{%
  \begin{tabular}{lcccccc}
    \hline
    \textbf{Prompts} & \multicolumn{2}{c}{\textbf{HumanEval}} & \multicolumn{2}{c}{\textbf{LiveCodeBench}} \\
    \cline{2-3} \cline{4-5}
                     & $pass$@$1$ & $pass$-$ratio$@$10$ & $pass$@$1$ & $pass$-$ratio$@$10$ \\
    \hline
    CGO-original               & \textbf{80.3} & 83.2 & \textbf{49.0} & 56.1  \\
    CGO-variant      & 79.27 & \textbf{83.9} & 48.5 & \textbf{56.19}  \\
    \hline
  \end{tabular}%
  }
  \caption{\label{Table7}
  Performance comparison of CGO and requirement-based prompting with the LLaMA3.1-70B-Instruct, evaluated on HumanEval and LiveCodeBench using $pass$@$1$ and $pass$-$ratio$@$10$.}
  
\end{table*}

\subsection{Wording Variations in CGO Prompting}

The CGO approach does not strictly require the use of the word ``objective'' in its prompts. For example, terms such as ``objective'' and ``requirement'' convey similar meanings within the context of coding. To evaluate whether minor wording differences could significantly influence the effectiveness of the CGO approach, we conducted an experiment.

In this experiment, we compared two variations of prompt wording used during the intermediate output generation stage:

\begin{itemize}
\item Original Prompt: \textit{``Write \textbf{objectives} to solve the problem''}
\item Variant Prompt: \textit{``Write \textbf{requirements} to solve the problem''}
\end{itemize}

We assessed the performance of code generation on the HumanEval and LiveCodeBench benchmarks. The experimental results are summarized in Table~\ref{Table7}.

The results reveal negligible differences between the two prompt variants, with variations within approximately 1\% in both \textit{pass@1} and \textit{pass-ratio@10} scores. These findings indicate that CGO’s effectiveness is consistently good regardless of minor semantic differences in prompt wording, allowing developers the flexibility to choose synonymous terms without compromising performance.

\subsection{Limitation}

Our study demonstrates the efficiency and effectiveness of CGO, but certain aspects remain unexplored:

\begin{itemize}
\item {\bf Model Selection}: The evaluation was conducted on five LLMs (LLaMA3-series, and GPT-3.5-Turbo), providing insights into CGO’s effectiveness but not covering the full range of recent models. Future research will extend experiments to Qwen, DeepSeek, and other emerging LLMs to assess CGO’s broader applicability.
\item {\bf Sampling Strategy}: We used greedy sampling for all experiments, ensuring deterministic outputs but not considering sampling-based approaches such as nucleus sampling. Since different sampling strategies can influence both output diversity and efficiency, future work will explore their impact on CGO.
\item \textbf{Objective Generation}: CGO generates objectives using a zero-shot approach, which may lead to variations in format, detail, or clarity across different instances. Employing methods such as few-shot prompting or structured guidance could improve consistency and standardization in the objectives produced. Additionally, systematic evaluations involving human or automated approaches (e.g., `LLM-as-judge'~\cite{gu2025surveyllmasajudge}) would further help assess the quality and appropriateness of generated objectives. 
\item \textbf{Evaluation Methodology}: Our evaluation primarily focused on functional correctness and performance metrics. Additional criteria such as readability, maintainability, and robustness against edge cases need to be considered.
\end{itemize}

Addressing these aspects will further refine CGO and enhance its adaptability to different models and inference settings.

\section{Conclusion}
This paper introduced Chain of Grounded Objectives (CGO), a concise goal-oriented reasoning approach that succinctly captures the objectives of a coding problem. CGO autonomously generates comment-style objectives, aligning well with programming language conventions while offering clear guidance without excessive verbosity.

Empirical evaluations on HumanEval, MBPP-sanitized, and their extended versions (HumanEval+ and MBPP+) demonstrate that CGO consistently achieves high accuracy, outperforming baseline prompting techniques in \textit{pass@1} and \textit{pass-ratio@10}. These results indicate that CGO enables LLMs to generate correct solutions without requiring intricate reasoning steps.

On LiveCodeBench, CGO also demonstrated competitive performance, particularly on medium and hard difficulty problems, underscoring its applicability to real-world programming tasks. In addition to accuracy, CGO exhibited the highest efficiency, often reducing the number of intermediate tokens generated by more than half compared to baseline prompts. This reduction minimizes inference costs while sustaining competitive performance, making CGO a cost-effective approach for LLM-based code generation.

Overall, CGO presents a practical and efficient approach to guiding LLMs in code generation. Future work could investigate applying CGO to specialized reasoning models such as DeepSeek-R1 or OpenAI o3 to further evaluate and enhance its effectiveness.
 


\bibliography{lipics-v2021-sample-article}
\clearpage
\appendix

\section{Experimental Details}
\label{sec:Expi_detail}
In our experiments, we employed LLaMA-3 and LLaMA-3.1 based models, and GPT-3.5-Turbo-16k-0106. Both objective generation and code generation were carried out using greedy sampling with a temperature of 0 and a top-p of 1, following prior work~\cite{han2024archcode, chen2022codet}.

The MBPP prompts were standardized to align with the HumanEval format. To ensure consistency, function signatures were extracted directly from the ground-truth code. Examples of the standardized formatting are presented in Table~\ref{humaneval_style_example_table}.

To increase the diversity of outputs, we generated \(n = 10\) objective samples per problem, which were subsequently used to guide the code generation process.

For the implementation of \textit{Chain-of-Thought (CoT)}~\cite{shao2023synthetic, zhang2022automatic} and \textit{Self-Planning} strategies, reasoning chains were constructed based on code examples provided in the original Self-Planning approach. Instead of directly reusing existing reasoning sequences, we manually curated new examples tailored to our experimental configuration.

Few-shot prompting was realized by randomly selecting example problems and their corresponding solutions from the same dataset. Each problem description was labeled as \textit{Input} and the corresponding solution as \textit{Output}, which were incorporated into the prompt as reference examples.

The execution of the generated code was configured to run in Docker sandbox environment.

All experiments were conducted using the Huggingface~\cite{jain2022hugging} and LangChain, LangGraph libraries. All resources are publicly available on GitHub.

\section{Evaluation Details}
\label{sec:eval_detail}
We evaluated the experimental results using \textit{pass@k} and \textit{pass-ratio@n} metrics.

\subsection{\textit{pass@k}}
The \textit{pass@k} metric assesses the performance of an LLM in code generation by evaluating the execution outcomes of the generated code. For a given coding problem, \textit{pass@k} indicates whether at least one of the \(k\) code solutions produced by the LLM successfully passes all associated test cases. A problem is deemed solved by the LLM if any of the \(k\) solutions satisfy this condition.

{\small
\begin{equation}
    \label{eq:passk}
        \textit{pass@k} := \underset{problems}{\mathbb{E}}\left[ 1 - \frac{\binom{n-c}{k}}{\binom{n}{k}} \right]
\end{equation}
}

\subsection{\textit{pass-ratio@n}}
For a given solution indexed by \(i\) (\(0 < i \leq n\)), \textit{pass-ratio}$_i$ is computed as the square of the ratio between the number of passed test cases for solution \(i\) and the total number of test cases. Squaring this value assigns greater weight to solutions with higher pass ratios, thereby emphasizing solutions with greater accuracy.

{\small
\begin{equation}
    \label{eq:passratio}
    \textit{pass-ratio}_i = \left( \frac{\text{Number of passed test cases } i}{\text{Total number of test cases}} \right)^2
\end{equation}
}

The \textit{pass-ratio@n} is calculated using the formula below.  It represents the average \textit{pass-ratio} of the $n$ generated codes.
\begin{equation}
    \label{eq:passration}
    \textit{pass-ratio@}n = \frac{\sum_{i=1}^n \textit{pass-ratio}_i}{n}
\end{equation}

\lstset{
    basicstyle=\ttfamily\small, 
    breaklines=true,            
    frame=none,                 
    showstringspaces=false      
}

\begin{figure*}[!h]
  \centering
  \caption*{\textbf{HumanEval Prompt Example}}
  \begin{lstlisting}
def separate_paren_groups(paren_string: str) -> List[str]:
    """ Input to this function is a string containing multiple groups of nested parentheses.
    >>> separate_paren_groups('( ) (( )) (( )( ))')
    ['()', '(())', '(()())']
    """
  \end{lstlisting}

  \vspace{0.5em}
  \noindent\rule{\textwidth}{0.4pt}
  \vspace{0.5em}

  \caption*{\textbf{MBPP-sanitized Prompt Example}}
  \begin{lstlisting}
def first_repeated_char(str1):
    '''Write a python function to find the first repeated character in a given string.'''
  \end{lstlisting}

  \caption{Examples of HumanEval-style Prompt (HumanEval, MBPP-sanitized)}
  \label{humaneval_style_example_table}
\end{figure*}

\begin{table*}[!t]
  \renewcommand{\arraystretch}{1.2}
  \centering
  \begin{tabular}{p{0.95\textwidth}}
    \hline
    \textbf{Problem Description Example} \\
    \hline
    \begin{minipage}{0.95\textwidth}
    \begin{lstlisting}
def separate_paren_groups(paren_string: str) -> List[str]:
    """ Input to this function is a string containing multiple groups of nested parentheses. Your goal is to
    separate those group into separate strings and return the list of those.
    Separate groups are balanced (each open brace is properly closed) and not nested within each other
    Ignore any spaces in the input string.
    >>> separate_paren_groups('( ) (( )) (( )( ))')
    ['()', '(())', '(()())']
    """
    \end{lstlisting}
    \end{minipage} \\
    \hline
    \textbf{Objectives Generation Prompt} \\
    Write objectives to solve the problem. \\
    \hline
    \textbf{Objectives Generation Examples} \\
    \begin{minipage}{0.95\textwidth}
    \begin{itemize}
      \item Separate groups of nested parentheses should be identified and extracted from the input string.
      \item Ignore any spaces in the input string.
      \item Ensure that each extracted group is balanced (each open brace is properly closed).
      \item Extracted groups should not be nested within each other.
    \end{itemize}
    \end{minipage} \\
    \hline
    \textbf{Generated Code} \\
    \begin{minipage}{0.95\textwidth}
    \begin{lstlisting}
def separate_paren_groups(paren_string: str) -> List[str]:
    result = []
    stack = []
    current_group = ""

    for char in paren_string:
        if char == '(':
            stack.append('(')
            current_group += '('
        elif char == ')':
            stack.pop()
            current_group += ')'
            if not stack:
                result.append(current_group)
                current_group = ""

    return result
    \end{lstlisting}
    \end{minipage} \\
    \hline
  \end{tabular}
  \caption{\label{cgo_example_table}
    Examples of Objective and Code Generation Using CGO
  }
\end{table*}

\begin{table*}[t]
  \renewcommand{\arraystretch}{1.1}
  \centering
  \begin{tabular}{p{0.95\textwidth}} 
    \hline
    \textbf{Problem Description Example} \\
    \begin{minipage}{0.95\textwidth}
    \begin{lstlisting}
def separate_paren_groups(paren_string: str) -> List[str]:
    """ Input to this function is a string containing multiple groups of nested parentheses. Your goal is to
    separate those group into separate strings and return the list of those.
    Separate groups are balanced (each open brace is properly closed) and not nested within each other
    Ignore any spaces in the input string.
    >>> separate_paren_groups('( ) (( )) (( )( ))')
    ['()', '(())', '(()())']
    """
    \end{lstlisting}
    \end{minipage} \\
    \hline
    \textbf{Few-shot Examples} \\
    \begin{minipage}{0.95\textwidth}
    \begin{lstlisting}
### Input
def minSubArraySum(nums):
    """
    Given an array of integers nums, find the minimum sum of any non-empty sub-array of nums. 
    Examples:
        minSubArraySum([2, 3, 4, 1, 2, 4]) == 1
        minSubArraySum([-1, -2, -3]) == -6
    """

###Output
def minSubArraySum(nums):
    max_sum = 0
    s = 0
    ...
    \end{lstlisting}
    \end{minipage} \\
    \hline
    \textbf{Few-shot Code Generation Prompt} \\
    Write python code for the problem. \\
    \hline
    \textbf{Generated Code} \\
    \begin{minipage}{0.95\textwidth}
    \begin{lstlisting}
groups = []
current_group = ""
for char in paren_string:
    if char == '(':
        current_group += char
    elif char == ')':
        current_group += char
        groups.append(current_group)
        current_group = ""
return groups
    \end{lstlisting}
    \end{minipage} \\
    \hline
  \end{tabular}
  \caption{\label{few-shot-example}
    Examples of Code Generation Using Few-shot
  }
\end{table*}

\begin{table*}[t]
  \renewcommand{\arraystretch}{1.1}
  \centering
  \begin{tabular}{p{0.95\textwidth}}
    \hline
    \textbf{Problem Description Example} \\
    \begin{minipage}{0.95\textwidth}
    \begin{lstlisting}
def separate_paren_groups(paren_string: str) -> List[str]:
    """ Input to this function is a string containing multiple groups of nested parentheses. Your goal is to
    separate those group into separate strings and return the list of those.
    Separate groups are balanced (each open brace is properly closed) and not nested within each other
    Ignore any spaces in the input string.
    >>> separate_paren_groups('( ) (( )) (( )( ))')
    ['()', '(())', '(()())']
    """
    \end{lstlisting}
    \end{minipage} \\
    \hline
    \textbf{CoT Generation Prompt} \\
    Generate detailed comments for this code. \\
    \hline
    \textbf{CoT Generation Examples} \\
    \begin{minipage}{0.95\textwidth}
    \begin{itemize}
      \item The input string contains groups of nested parentheses separated by spaces.
      \item Split the input string by spaces to handle each group individually.
      \item Remove any extra spaces to ensure proper processing.
      \item Assume each group is balanced and valid.
      \item Iterate through groups, filtering out empty or invalid ones.
      \item Collect remaining valid groups into a list.
      \item Return the filtered list.
      \item This method assumes no nested validation beyond splitting.
    \end{itemize}
    \end{minipage} \\
    \hline
    \textbf{Generated Code} \\
    \begin{minipage}{0.95\textwidth}
    \begin{lstlisting}
groups = paren_string.split(' ') 
filtered_groups = [group.strip() for group in groups if group.strip()]  
return filtered_groups

print(separate_paren_groups('( ) (( )) (( )( ))'))  # Output: ['()', '(())', '(()())']
    \end{lstlisting}
    \end{minipage} \\
    \hline
  \end{tabular}
  \caption{\label{cot-example}
    Examples of Code Generation Using CodeCoT
  }
\end{table*}

\begin{table*}[t]
  \renewcommand{\arraystretch}{1.1}
  \centering
  \begin{tabular}{p{0.95\textwidth}} 
    \hline
    \textbf{Problem Description Example} \\
    \begin{minipage}{0.95\textwidth}
    \begin{lstlisting}
def separate_paren_groups(paren_string: str) -> List[str]:
    """ Input to this function is a string containing multiple groups of nested parentheses. Your goal is to
    separate those group into separate strings and return the list of those.
    Separate groups are balanced (each open brace is properly closed) and not nested within each other
    Ignore any spaces in the input string.
    >>> separate_paren_groups('( ) (( )) (( )( ))')
    ['()', '(())', '(()())']
    """
    \end{lstlisting}
    \end{minipage} \\
    \hline
    \textbf{Plan Generation Prompt} \\
    Write plan for the problem. \\
    \hline
    \textbf{Plan Generation Examples} \\
    \begin{minipage}{0.95\textwidth}
    \begin{enumerate}[(1)]
      \item Preprocess the input string:
      \begin{enumerate}[(a)]
        \item Remove all spaces to simplify processing.
      \end{enumerate}
    
      \item Initialize variables:
      \begin{enumerate}[(a)]
        \item List \texttt{groups} to store the final balanced groups.
        \item String \texttt{temp} to construct the current group.
        \item Integer \texttt{balance} to track open and close parentheses.
      \end{enumerate}
    
      \item Iterate through the cleaned string:
      \begin{enumerate}[(a)]
        \item Add the character to \texttt{temp}.
        \item Adjust \texttt{balance} based on whether the character is \texttt{(} or \texttt{)}.
        \item If \texttt{balance == 0}, add \texttt{temp} to \texttt{groups} and reset \texttt{temp}.
      \end{enumerate}
    
      \item Return the list \texttt{groups} containing all balanced groups.
    \end{enumerate}
    \end{minipage} \\
    \hline
    \textbf{Generated Code} \\
    \begin{minipage}{0.95\textwidth}
    \begin{lstlisting}
cleaned_string = paren_string.replace(" ", "")
groups = []
temp = ""
balance = 0
for char in cleaned_string:
    temp += char
    if char == '(':
        balance += 1
    elif char == ')':
        balance -= 1
        if balance == 0:
            groups.append(temp)
            temp = ""

return groups
    \end{lstlisting}
    \end{minipage} \\
    \hline
  \end{tabular}
  \caption{\label{plan-example}
    Examples of Code Generation Using Self-Planning
  }
\end{table*}

\begin{table*}[t]
  \renewcommand{\arraystretch}{1.1}
  \centering
  \begin{tabular}{p{0.95\textwidth}} 
    \hline
    \textbf{Problem Description Example} \\
    \begin{minipage}{0.95\textwidth}
    \begin{lstlisting}
def separate_paren_groups(paren_string: str) -> List[str]:
    """ Input to this function is a string containing multiple groups of nested parentheses. Your goal is to
    separate those group into separate strings and return the list of those.
    Separate groups are balanced (each open brace is properly closed) and not nested within each other
    Ignore any spaces in the input string.
    >>> separate_paren_groups('( ) (( )) (( )( ))')
    ['()', '(())', '(()())']
    """
    \end{lstlisting}
    \end{minipage} \\
    \hline

    \textbf{Pseudo-code} \\
    \begin{minipage}{0.95\textwidth}
    \begin{lstlisting}
Function separate_paren_groups(input_string):
    Strip spaces from input_string
    Initialize result as empty list
    Initialize current_group as empty string
    Set balance = 0
    For each char in input_string:
        Append char to current_group
        If char is '(':
            increment balance
        If char is ')':
            decrement balance
        If balance == 0:
            Append current_group to result
            Reset current_group to empty string
    Return result
    \end{lstlisting}
    \end{minipage} \\
    \hline

    \textbf{Generated Code} \\
    \begin{minipage}{0.95\textwidth}
    \begin{lstlisting}
cleaned_string = paren_string.replace(" ", "")
groups = []
temp = ""
balance = 0
for char in cleaned_string:
    temp += char
    if char == '(':
        balance += 1
    elif char == ')':
        balance -= 1
        if balance == 0:
            groups.append(temp)
            temp = ""
return groups
    \end{lstlisting}
    \end{minipage} \\
    \hline
  \end{tabular}
  \caption{\label{pseudo-example}
    Examples of Code Generation Using Pseudo-Code
  }
\end{table*}

\end{document}